\documentclass[11pt,a4paper]{article}
\usepackage[hyperref]{acl2018}
\usepackage[pdftex]{graphicx} 
\usepackage{graphicx}
\usepackage{times}
\usepackage{latexsym}
\usepackage{array}
\usepackage{amssymb}
\usepackage{ascmac}
\usepackage{amssymb}
\usepackage{algorithm}
\usepackage{mathtools}
\usepackage{dcolumn}

\def \R{{\Bbb R}}

\DeclareMathOperator*{\argmax}{arg\,max}

\usepackage{url}

\aclfinalcopy 
\def\aclpaperid{94} 

\newcolumntype{R}{>{\centering\arraybackslash}p{2.5em}}
\newcolumntype{Y}{>{\centering\arraybackslash}p{3.5em}}
\newcolumntype{Z}{>{\centering\arraybackslash}p{4.5em}}
\newcolumntype{X}{>{\centering\arraybackslash}p{5em}}
\newcolumntype{Q}{>{\centering\arraybackslash}p{2.3em}}

\title{Subword Regularization: Improving Neural Network Translation Models with Multiple Subword Candidates}

\author{Taku Kudo \\
  Google Inc.\\
  {\tt taku@google.com} \\
}

\date{}

\begin{document}
\maketitle
 \begin{abstract}
  Subword units are an effective way to alleviate the open vocabulary
  problems in neural machine translation (NMT). While sentences are
  usually converted into unique subword sequences, subword segmentation
  is potentially ambiguous and multiple segmentations are possible even
  with the same vocabulary. The question addressed in this paper is
  whether it is possible to harness the segmentation ambiguity as a
  noise to improve the robustness of NMT. We present a simple
  regularization method, subword regularization, which trains the model
  with multiple subword segmentations probabilistically sampled during
  training.  In addition, for better subword sampling, we propose a new
  subword segmentation algorithm based on a unigram language model.  We
  experiment with multiple corpora and report consistent improvements
  especially on low resource and out-of-domain settings.
 \end{abstract}

\section{Introduction}
Neural Machine Translation (NMT) models
\cite{bahdanau2014neural,luong2015effective,wu2016google,ashish2017google}
often operate with fixed word vocabularies, as their training and
inference depend heavily on the vocabulary size.  However, limiting
vocabulary size increases the amount of unknown words, which makes the
translation inaccurate especially in an open vocabulary setting.

  \begin{table}[!ht]
   \renewcommand{\arraystretch}{0.95}
  \begin{center}
    \begin{tabular}{l|l}
      \hline
      \small{Subwords} \tiny{(\_ means spaces)}& \small{Vocabulary id sequence} \\
      \hline
      \_Hell/o/\_world & 13586 137 255 \\
      \_H/ello/\_world & 320 7363 255  \\
      \_He/llo/\_world & 579 10115 255  \\
      \_/He/l/l/o/\_world & 7 18085 356 356 137 255 \\
      \_H/el/l/o/\_/world & 320 585 356 137 7 12295 \\
      \hline
   \end{tabular}
  \end{center}
 \vspace*{-5mm}
  \caption{Multiple subword sequences encoding the same sentence ``Hello
 World''}\label{subword}
 \vspace*{-4mm}
  \end{table}

A common approach for dealing with the open vocabulary issue is to
break up rare words into subword units
\cite{schuster2012japanese,chitnis2015variable,sennrichneural,wu2016google}.
Byte-Pair-Encoding (BPE) \cite{sennrichneural} is a de facto standard subword
segmentation algorithm applied to many NMT systems and achieving top
translation quality in several shared tasks
\cite{denkowski2017stronger,nakazawa2017overview}.  BPE segmentation
gives a good balance between the vocabulary size and the decoding
efficiency, and also sidesteps the need for a special treatment of
unknown words.

BPE encodes a sentence into a unique subword sequence.  However, a
sentence can be represented in multiple subword sequences even with the
same vocabulary. Table \ref{subword} illustrates an example. While these
sequences encode the same input ``Hello World'', NMT handles them as
completely different inputs. This observation becomes more apparent when
converting subword sequences into id sequences (right column in Table
\ref{subword}). These variants can be viewed as a spurious ambiguity,
which might not always be resolved in decoding process.  At training
time of NMT, multiple segmentation candidates will make the model robust
to noise and segmentation errors, as they can indirectly help the
model to learn the compositionality of words, e.g., ``books'' can be
decomposed into ``book'' + ``s''.

In this study, we propose a new regularization method for
open-vocabulary NMT, called \textbf{subword regularization}, which
employs multiple subword segmentations to make the NMT model accurate
and robust. Subword regularization consists of the following two
sub-contributions:

\begin{itemize}
 \item We propose a simple NMT training algorithm to integrate multiple
       segmentation candidates. Our approach is implemented as an on-the-fly
       data sampling, which is not specific to NMT architecture.  Subword
       regularization can be applied to any NMT system without changing the
       model structure.
 \item We also propose a new subword segmentation algorithm based on a language
       model, which provides multiple segmentations with probabilities.
       The language model allows to emulate the noise generated during the
       segmentation of actual data.
\end{itemize}

Empirical experiments using multiple corpora with different sizes and
languages show that subword regularization achieves significant
improvements over the method using a single subword sequence. In
addition, through experiments with out-of-domain corpora, we show that
subword regularization improves the robustness of the NMT model.

\section{Neural Machine Translation with multiple subword segmentations}
\subsection{NMT training with on-the-fly subword sampling}
Given a source sentence $X$ and a target sentence $Y$, let $\mathbf{x} =
(x_1,\ldots,x_M)$ and $\mathbf{y}=(y_1,\ldots,y_N)$ be the corresponding
subword sequences segmented with an underlying subword segmenter, e.g.,
BPE.  NMT models the translation probability $P(Y|X) =
P(\mathbf{y}|\mathbf{x})$ as a target language sequence model that
generates target subword $y_n$ conditioning on the target history
$y_{<n}$ and source input sequence $\mathbf{x}$:
\begin{eqnarray}
  P(\mathbf{y}|\mathbf{x};\mathbf{\theta}) = \prod_{n=1}^{N}P(y_n|\mathbf{x}, y_{<n};\mathbf{\theta}),\label{nmt}
\end{eqnarray}
where $\theta$ is a set of model parameters. A common choice to predict
the subword $y_n$ is to use a recurrent neural network (RNN)
architecture.  However, note that subword regularization is not
specific to this architecture and can be applicable to other NMT
architectures without RNN, e.g., \cite{ashish2017google,gehring2017convolutional}.

NMT is trained using the standard maximum likelihood estimation, i.e.,
maximizing the log-likelihood $\mathcal{L}(\mathbf{\theta})$ of a given
parallel corpus $D=\{\langle X^{(s)}, Y^{(s)}\rangle\}_{s=1}^{|D|} =
\{\langle \mathbf{x}^{(s)}, \mathbf{y}^{(s)}\rangle\}_{s=1}^{|D|}$,
\begin{eqnarray}
  \mathbf{\theta}_{MLE} &=& \argmax_{\mathbf{\theta}} \mathcal{L}(\mathbf{\theta}) \nonumber \\
  where,\,\, \mathcal{L}(\mathbf{\theta}) &=&
  \sum_{s=1}^{|D|} \log P(\mathbf{y}^{(s)}|\mathbf{x}^{(s)}; \mathbf{\theta}).
  \label{nmt_train}
\end{eqnarray}

We here assume that the source and target sentences $X$ and $Y$ can be
segmented into multiple subword sequences with the segmentation
probabilities $P(\mathbf{x}|X)$ and $P(\mathbf{y}|Y)$ respectively. In
subword regularization, we optimize the parameter set $\theta$ with the
marginalized likelihood as (\ref{nmt_train_marginal}).
\begin{eqnarray}
  \mathcal{L}_{marginal}(\mathbf{\theta}) =
  \sum_{s=1}^{|D|}
  \mathbb{E}_{\substack{\mathbf{x} \sim P(\mathbf{x}|X^{(s)}) \\
                  \mathbf{y} \sim P(\mathbf{y}|Y^{(s)})}}[
  \log P(\mathbf{y}|\mathbf{x}; \mathbf{\theta})] \label{nmt_train_marginal}
\end{eqnarray}
Exact optimization of (\ref{nmt_train_marginal}) is not feasible as the
number of possible segmentations increases exponentially with respect to
the sentence length. We approximate (\ref{nmt_train_marginal}) with
finite $k$ sequences sampled from $P(\mathbf{x}|X)$ and
$P(\mathbf{y}|Y)$ respectively.
\begin{eqnarray}
  \begin{split}
    \mathcal{L}_{marginal}(\mathbf{\theta})& \cong \frac{1}{k^2}\sum_{s=1}^{|D|}
  \sum_{i=1}^{k}
  \sum_{j=1}^{k}
  \log P(\mathbf{y}_j|\mathbf{x}_i; \mathbf{\theta}) \\
  \mathbf{x}_i& \sim P(\mathbf{x}|X^{(s)}),\,\,\,\,\mathbf{y}_j \sim P(\mathbf{y}|Y^{(s)}).
  \label{nmt_train_marginal2}
  \end{split}
\end{eqnarray}
For the sake of simplicity, we use $k=1$.  Training of NMT usually uses
an online training for efficiency, in which the parameter $\theta$ is
iteratively optimized with respect to the smaller subset of $D$
(mini-batch). When we have a sufficient number of iterations, subword
sampling is executed via the data sampling of online training, which
yields a good approximation of (\ref{nmt_train_marginal}) even if $k=1$.
It should be noted, however, that the subword sequence is sampled
on-the-fly for each parameter update.

\subsection{Decoding}
In the decoding of NMT, we only have a raw source sentence $X$.  A
straightforward approach for decoding is to translate from the best
segmentation $\mathbf{x}^{*}$ that maximizes the probability
$P(\mathbf{x}|X)$, i.e., $\mathbf{x}^{*} = argmax_\mathbf{x}
P(\mathbf{x}|X)$.  Additionally, we can use the $n$-best segmentations of
$P(\mathbf{x}|X)$ to incorporate multiple segmentation candidates. More
specifically, given $n$-best segmentations
$(\mathbf{x}_1,\ldots,\mathbf{x}_n)$, we choose the best translation
$\mathbf{y}^*$ that maximizes the following score.
\begin{eqnarray}
  score(\mathbf{x}, \mathbf{y}) = \log P(\mathbf{y}|\mathbf{x})/|\mathbf{y}|^\lambda,
\end{eqnarray}
where $|\mathbf{y}|$ is the number of subwords in  $\mathbf{y}$ and
$\lambda \in \R^{+}$ is the parameter to penalize shorter sentences.
$\lambda$ is optimized with the development data.

In this paper, we call these two algorithms \textbf{one-best decoding} and
\textbf{$n$-best decoding} respectively.

\section{Subword segmentations with language model}
\subsection{Byte-Pair-Encoding (BPE)}
Byte-Pair-Encoding (BPE) \cite{sennrichneural, schuster2012japanese} is
a subword segmentation algorithm widely used in many NMT
systems\footnote{Strictly speaking, wordpiece model
\cite{schuster2012japanese} is different from BPE. We consider wordpiece
as a variant of BPE, as it also uses an incremental vocabulary
generation with a different loss function.}.  BPE first splits the whole
sentence into individual characters. The most
frequent\footnote{Wordpiece model uses a likelihood instead of
frequency.} adjacent pairs of characters are then consecutively merged
until reaching a desired vocabulary size. Subword segmentation is
performed by applying the same merge operations to the test sentence.

An advantage of BPE segmentation is that it can effectively balance
the vocabulary size and the step size (the number of tokens required
to encode the sentence). BPE trains the merged operations only with a
frequency of characters. Frequent substrings will be joined early,
resulting in common words remaining as one unique symbol. Words
consisting of rare character combinations will be split into smaller
units, e.g., substrings or characters.  Therefore, only with a
small fixed size of vocabulary (usually 16k to 32k), the number of
required symbols to encode a sentence will not significantly increase,
which is an important feature for an efficient decoding.

One downside is, however, that BPE is based on a greedy and
deterministic symbol replacement, which can not provide multiple segmentations
with probabilities. It is not trivial to apply BPE to
the subword regularization that depends on segmentation probabilities
$P(\mathbf{x}|X)$.

\subsection{Unigram language model}
In this paper, we propose a new subword segmentation algorithm based on
a unigram language model, which is capable of outputing multiple subword
segmentations with probabilities.  The unigram language model makes an
assumption that each subword occurs independently, and consequently, the
probability of a subword sequence $\mathbf{x} = (x_1,\ldots,x_M)$ is
formulated as the product of the subword occurrence probabilities
$p(x_i)$\footnote{Target sequence $\mathbf{y} = (y_1,\ldots,y_N)$ can
also be modeled similarly.}:
\begin{eqnarray}
  P(\mathbf{x}) = \prod_{i=1}^{M} p(x_i), \\
  \forall i\,\, x_i \in \mathcal{V},\,\,\,
  \sum_{x \in \mathcal{V}} p(x) = 1, \nonumber \label{unigram}
\end{eqnarray}
where $\mathcal{V}$ is a pre-determined vocabulary.  The most probable
segmentation $\mathbf{x}^*$ for the input sentence $X$ is then given by
\begin{eqnarray}
  \mathbf{x}^{*} = \argmax_{\mathbf{x} \in \mathcal{S}(X)} P(\mathbf{x}), \label{viterbi}
\end{eqnarray}
where $\mathcal{S}(X)$ is a set of segmentation candidates built from
the input sentence $X$.  $\mathbf{x}^*$ is obtained with the Viterbi
algorithm \cite{viterbi67}.

If the vocabulary $\mathcal{V}$ is given, subword occurrence
probabilities $p(x_i)$ are estimated via the EM algorithm that maximizes
the following marginal likelihood $\mathcal{L}$ assuming that $p(x_i)$
are hidden variables.
\begin{eqnarray*}
  \mathcal{L} = \sum_{s=1}^{|D|} \log(P(X^{(s)})) = \sum_{s=1}^{|D|}
   \log\Bigl(\sum_{\mathbf{x} \in \mathcal{S}(X^{(s)}) } P(\mathbf{x})\Bigr)
\label{EM}
\end{eqnarray*}

In the real setting, however, the vocabulary set $\mathcal{V}$ is also
unknown. Because the joint optimization of vocabulary set and their
occurrence probabilities is intractable, we here seek to find them with
the following iterative algorithm.
\begin{enumerate}
\item Heuristically make a reasonably big seed vocabulary from the training corpus.
\item Repeat the following steps until $|\mathcal{V}|$ reaches a desired
vocabulary size.
  \begin{enumerate}
  \item Fixing the set of vocabulary, optimize $p(x)$ with the EM
        algorithm.
  \item Compute the $loss_i$ for each subword $x_i$, where $loss_i$ represents
        how likely the likelihood $\mathcal{L}$ is reduced when the subword
        $x_i$ is removed from the current vocabulary.
  \item Sort the symbols by $loss_i$ and keep top $\eta$ \% of subwords
        ($\eta$ is 80, for example).  Note that we always keep the
        subwords consisting of a single character to avoid out-of-vocabulary.
  \end{enumerate}
\end{enumerate}
There are several ways to prepare the seed vocabulary. The natural
choice is to use the union of all characters and the most frequent
substrings in the corpus\footnote{It is also possible to run BPE with
a sufficient number of merge operations.}.  Frequent substrings can
be enumerated in $O(T)$ time and $O(20T)$ space with the Enhanced
Suffix Array algorithm \cite{nong2009linear}, where $T$ is the size of
the corpus. Similar to \cite{sennrichneural}, we do not consider subwords that
cross word boundaries.

As the final vocabulary $\mathcal{V}$ contains all individual characters
in the corpus, character-based segmentation is also included in the set
of segmentation candidates $\mathcal{S}(X)$. In other words, subword
segmentation with the unigram language model can be seen as a
probabilsitic mixture of characters, subwords and word segmentations.

\subsection{Subword sampling}
Subword regularization samples one subword segmentation from the
distribution $P(\mathbf{x}|X)$ for each parameter update.  A
straightforward approach for an approximate sampling is to use the
$l$-best segmentations.  More specifically, we first obtain $l$-best
segmentations according to the probability $P(\mathbf{x}|X)$.  $l$-best
search is performed in linear time with the Forward-DP Backward-A*
algorithm \cite{nagata1994stochastic}. One segmentation $\mathbf{x}_i$
is then sampled from the multinomial distribution $P(\mathbf{x}_i|X)
\cong P(\mathbf{x}_i)^{\alpha}/\sum_{i=1}^{l}
P(\mathbf{x}_{i})^{\alpha}$, where $\alpha \in \R^{+}$ is the
hyperparameter to control the smoothness of the distribution.  A
smaller $\alpha$ leads to sample $\mathbf{x}_i$ from a more uniform
distribution. A larger $\alpha$ tends to select the Viterbi
segmentation.

Setting $l \to \infty$, in theory, allows to take all possible
segmentations into account. However, it is not feasible to increase $l$
explicitly as the number of candidates increases exponentially with
respect to the sentence length.  In order to exactly sample from all
possible segmentations, we use the Forward-Filtering and
Backward-Sampling algorithm (FFBS) \cite{scott2002bayesian}, a variant
of the dynamic programming originally introduced by Bayesian hidden
Markov model training. In FFBS, all segmentation candidates are
represented in a compact lattice structure, where each node denotes a
subword. In the first pass, FFBS computes a set of forward
probabilities for all subwords in the lattice, which provide the
probability of ending up in any particular subword $w$. In the second
pass, traversing the nodes in the lattice from the end of the sentence
to the beginning of the sentence, subwords are recursively sampled for
each branch according to the forward probabilities.

\subsection{BPE vs. Unigram language model}
BPE was originally introduced in the data compression literature
\cite{gage1994}. BPE is a variant of dictionary (substitution) encoder
that incrementally finds a set of symbols such that the total number of
symbols for encoding the text is minimized.  On the other hand, the
unigram language model is reformulated as an entropy encoder that
minimizes the total code length for the text. According to Shannon's
coding theorem, the optimal code length for a symbol $s$ is $-\log p_s$,
where $p_s $ is the occurrence probability of $s$. This is essentially
the same as the segmentation strategy of the unigram language model
described as (\ref{viterbi}).

BPE and the unigram language model share the same idea that they encode
a text using fewer bits with a certain data compression principle
(dictionary vs. entropy).  Therefore, we expect to see the same benefit
as BPE with the unigram language model.  However, the unigram language
model is more flexible as it is based on a probabilistic language model
and can output multiple segmentations with their probabilities, which is
an essential requirement for subword regularization.

\section{Related Work}
Regularization by noise is a well studied technique in deep neural
networks.  A well-known example is dropout
\cite{srivastava2014dropout}, which randomly turns off a subset of
hidden units during training.
Dropout is analyzed as an ensemble
training, where many different models are trained on different subsets of the
data. Subword regularization trains the model on different data inputs randomly
sampled from the original input sentences, and thus is regarded as a variant of
ensemble training.

The idea of noise injection has previously been used in the context of
Denoising Auto-Encoders (DAEs) \cite{vincent2008extracting}, where
noise is added to the inputs and the model is trained to reconstruct
the original inputs. There are a couple of studies that employ DAEs in
natural language processing.

\cite{guillaume2017,mikel2017}
independently propose DAEs in the context of sequence-to-sequence
learning, where they randomly alter the word order of the input
sentence and the model is trained to reconstruct the original
sentence.  Their technique is applied to an unsupervised machine
translation to make the encoder truly learn the compositionality of
input sentences.

Word dropout \cite{iyyer2015deep} is a simple
approach for a bag-of-words representation, in which the embedding of
a certain word sequence is simply calculated by averaging the word
embeddings. Word dropout randomly drops words from the bag before
averaging word embeddings, and consequently can see $2^{|X|}$
different token sequences for each input $X$.

\cite{belinkov2017}
explore the training of character-based NMT with a synthetic noise
that randomly changes the order of characters in a word.
\cite{xie2017} also proposes a robust RNN language model that
interpolates random unigram language model.

The basic idea and
motivation behind subword regularization are similar to those of
previous work. In order to increase the robustness, they inject noise
to input sentences by randomly changing the internal representation of
sentences. However, these previous approaches often depend on
heuristics to generate synthetic noises, which do not always reflect
the real noises on training and inference. In addition, these approaches
can only be applied to source sentences (encoder), as they irreversibly rewrite
the surface of sentences.
Subword regularization, on the other hand, generates synthetic subword
sequences with an underlying language model to better emulate the
noises and segmentation errors. As subword regularization is based on
an invertible conversion, we can safely apply it both to source and
target sentences.

Subword regularization can also be viewed as a data augmentation. In
subword regularization, an input sentence is converted into multiple
invariant sequences, which is similar to the data augmentation
for image classification tasks, for example, random flipping,
distorting, or cropping.

There are several studies focusing on segmentation ambiguities in
language modeling.  Latent Sequence Decompositions (LSDs)
\cite{chan2016latent} learns the mapping from the input and the output
by marginalizing over all possible segmentations. LSDs and subword
regularization do not assume a predetermined segmentation for a
sentence, and take multiple segmentations by a similar marginalization
technique. The difference is that subword regularization injects the
multiple segmentations with a separate language model through an
on-the-fly subword sampling.  This approach makes the model simple and
independent from NMT architectures.

Lattice-to-sequence models \cite{su2017lattice,sperber2017neural} are
natural extension of sequence-to-sequence models, which represent
inputs uncertainty through lattices. Lattice is encoded with a variant
of TreeLSTM \cite{tai2015improved}, which requires changing the model
architecture. In addition, while subword regularization is applied
both to source and target sentences, lattice-to-sequence models do not
handle target side ambiguities.

A mixed word/character model \cite{wu2016google} addresses the
out-of-vocabulary problem with a fixed vocabulary. In this model,
out-of-vocabulary words are not collapsed into a single UNK symbol, but
converted into the sequence of characters with special prefixes
representing the positions in the word. Similar to BPE, this model also
encodes a sentence into a unique fixed sequence, thus multiple
segmentations are not taken into account.

\section{Experiments}
\subsection{Setting}
We conducted experiments using multiple corpora with different sizes and
languages.  Table \ref{corpus} summarizes the evaluation data we used
\footnote{IWSLT15: \url{http://workshop2015.iwslt.org/}}
\footnote{IWSLT17: \url{http://workshop2017.iwslt.org/}} \footnote{KFTT:
\url{http://www.phontron.com/kftt/}} \footnote{ASPEC:
\url{http://lotus.kuee.kyoto-u.ac.jp/ASPEC/}} \footnote{WMT14:
\url{http://statmt.org/wmt14/}} \footnote{WMT14(en$\leftrightarrow$de)
uses the same setting as \cite{wu2016google}.}.  IWSLT15/17 and KFTT are
relatively small corpora, which include a wider spectrum of languages
with different linguistic properties. They can evaluate the
language-agnostic property of subword regularization. ASPEC and WMT14
(en$\leftrightarrow$de) are medium-sized corpora.  WMT14
(en$\leftrightarrow$cs) is a rather big corpus consisting of more than 10M
parallel sentences.

We used GNMT \cite{wu2016google} as the implementation of the NMT system
for all experiments.  We generally followed the settings and training
procedure described in \cite{wu2016google}, however, we changed the
settings according to the corpus size.  Table \ref{corpus} shows the
hyperparameters we used in each experiment. As common settings, we set
the dropout probability to be 0.2. For parameter estimation, we used a
combination of Adam \cite{kingma6980method} and SGD algorithms. Both
length normalization and converge penalty parameters are set to 0.2 (see
section 7 in \cite{wu2016google}). We set the decoding beam size to
4. 

The data was preprocessed with Moses tokenizer before training subword
models. It should be noted, however, that Chinese and Japanese have no
explicit word boundaries and Moses tokenizer does not segment sentences
into words, and hence subword segmentations are trained almost from
unsegmented raw sentences in these languages.

We used the case sensitive BLEU score \cite{papineni2002bleu} as an
evaluation metric.  As the output sentences are not segmented in
Chinese and Japanese, we segment them with characters and
KyTea\footnote{\url{http://www.phontron.com/kytea}} for
Chinese and Japanese respectively before calculating BLEU scores.

BPE segmentation is used as a baseline system.  We evaluate three test
systems with different sampling strategies: (1) Unigram language
model-based subword segmentation without subword regularization
($l\!=\!1$), (2) with subword regularization
$(l\!=\!64,\,\,\alpha\!=\!0.1$) and (3)
$(l\!=\!\infty,\,\,\alpha\!=\!0.2/0.5)\,\,$ 0.2: IWSLT, 0.5: others.
These sampling parameters were
determined with preliminary experiments. $l\!=\!1$ is aimed at a pure
comparison between BPE and the unigram language model. In addition, we
compare one-best decoding and $n$-best decoding (See section 2.2). Because
BPE is not able to provide multiple segmentations, we only evaluate
one-best decoding for BPE.  Consequently, we compare 7 systems (1 + 3
$\times$ 2) for each language pair.

\begin{table*}[t]
  \renewcommand{\arraystretch}{0.9}
 \begin{center}
  \begin{tabular}[c]{l|c|R|R|R|X|X|X}
   \hline
   & & \multicolumn{3}{c|}{Size of sentences} & \multicolumn{3}{c}{Parameters} \\
   \cline{3-8}
   {\small Corpus} & {\small\shortstack{Language\\pair}} & train & dev & test &
   {\scriptsize \shortstack{{}\\\#vocab\\(Enc/Dec shared)}} &
       {\scriptsize \shortstack{{}\\\#dim of LSTM，\\embedding}} &
       {\scriptsize \shortstack{{}\\\#layers of LSTM \\(Enc+Dec)}} \\
   \hline
   {\small IWSLT15} & en $\leftrightarrow$ vi & 133k & 1553 & 1268 & 16k & 512 & 2+2\\
   & en $\leftrightarrow$ zh & 209k & 887 & 1261 & 16k & 512  & 2+2\\
   {\small IWSLT17} & en $\leftrightarrow$ fr & 232k & 890 & 1210 & 16k & 512 & 2+2\\
   & en $\leftrightarrow$ ar & 231k & 888 & 1205 & 16k & 512 & 2+2\\
   {\small KFTT} &  en $\leftrightarrow$ ja & 440k & 1166 & 1160 & 8k & 512 & 6+6\\
   {\small ASPEC} &  en $\leftrightarrow$ ja & 2M & 1790 & 1812 & 16k & 512 & 6+6\\
   {\small WMT14} &  en $\leftrightarrow$ de & 4.5M& 3000 & 3003 & 32k & 1024 & 8+8\\
   &  en $\leftrightarrow$ cs & 15M& 3000 & 3003 & 32k & 1024 & 8+8\\
   \hline
  \end{tabular}
 \end{center}
 \vspace*{-4mm}
 \caption{Details of evaluation data set}\label{corpus}
\end{table*}

\subsection{Main Results}
Table \ref{result} shows the translation experiment results.

First, as can be seen in the table, BPE and unigram language model
without subword regularization ($l=1$) show almost comparable BLEU scores.
This is not surprising, given that
both BPE and the unigram language model are based on data compression
algorithms.

We can see that subword regularization ($l>1$) boosted BLEU scores quite
impressively (+1 to 2 points) in all language pairs except for WMT14 (en$\rightarrow$cs) dataset.
The gains are larger especially in lower resource settings (IWSLT and
KFTT). It can be considered that the positive effects of data
augmentation with subword regularization worked better in lower
resource settings, which is a common property of other regularization
techniques.

As for the sampling algorithm, $(l\!=\!\infty\,\,\alpha\!=\!0.2/0.5)$ slightly
outperforms $(l\!=\!64,\,\,\alpha\!=\!0.1)$ on IWSLT corpus, but they show
almost comparable results on larger data set. Detailed analysis is described in Section 5.5.

On top of the gains with subword regularization, $n$-best decoding yields
further improvements in many language pairs. However, we should note that the subword
regularization is mandatory for $n$-best decoding and the BLEU score is
degraded in some language pairs without subword regularization
($l=1$). This result indicates that the decoder is more confused for
multiple segmentations when they are not explored at training time.

\begin{table*}[t]
  \renewcommand{\arraystretch}{0.88}
  \begin{center}
    \begin{tabular}{l|c|Y|Y|Y|Y|Y|Y|Y}
      \hline
      &        &   &
     \multicolumn{3}{c|}{{\small Proposed (one-best decoding)}} &
     \multicolumn{3}{c}{{\small Proposed ($n$-best decoding, $n\!=\!64$)}} \\
      \cline{4-9}
     {\small Corpus} & {\small\shortstack{Language\\pair}} & {\small\shortstack{baseline\\(BPE)}} &
      {\small $l=1$} & {\small\shortstack{$l=64$\\$\alpha=0.1$}} & {\small \shortstack{{}\\$l=\infty$\\\hspace*{-1mm}$\alpha\!=\!0.2/0.5$}} 
      & {\small $l=1$} & {\small \shortstack{$l=64$\\$\alpha=0.1$}} & {\small \shortstack{$l=\infty$\\\hspace*{-1mm}$\alpha\!=\!0.2/0.5$}} \\
      \hline
\small{IWSLT15} & en $\to$ vi & 25.61 & 25.49\phantom{*} & 27.68* & 27.71* & 25.33\phantom{*} & 28.18* & 28.48* \\
    & vi $\to$ en & 22.48 & 22.32\phantom{*} & 24.73* & 26.15* & 22.04\phantom{*} & 24.66* & 26.31* \\
 & en $\to$ zh & 16.70 & 16.90\phantom{*} & 19.36* & 20.33* & 16.73\phantom{*} & 20.14* & 21.30* \\
    & zh $\to$ en & 15.76 & 15.88\phantom{*} & 17.79* & 16.95* & 16.23\phantom{*} & 17.75* & 17.29* \\
\hline
\small{IWSLT17} & en $\to$ fr & 35.53 & 35.39\phantom{*} & 36.70* & 36.36* & 35.16\phantom{*} & 37.60* & 37.01* \\
                & fr $\to$ en & 33.81 & 33.74\phantom{*} & 35.57* & 35.54* & 33.69\phantom{*} & 36.07* & 36.06* \\
 & en $\to$ ar & 13.01 & 13.04\phantom{*} & 14.92* & 15.55* & 12.29\phantom{*} & 14.90* & 15.36* \\
    & ar $\to$ en & 25.98 & 27.09* & 28.47* & 29.22* & 27.08* & 29.05* & 29.29* \\
\hline
\small{KFTT} & en $\to$ ja & 27.85 & 28.92* & 30.37* & 30.01* & 28.55* & 31.46* & 31.43* \\
& ja $\to$ en & 21.37 & 21.46\phantom{*} & 22.33* & 22.04* & 21.37\phantom{*} & 22.47* & 22.64* \\
\hline
\small{ASPEC} & en $\to$ ja & 40.62 & 40.66\phantom{*} & 41.24* & 41.23* & 40.86\phantom{*} & 41.55* & 41.87* \\
& ja $\to$ en & 26.51 & 26.76\phantom{*} & 27.08* & 27.14* & 27.49* & 27.75* & 27.89* \\
\hline
\small{WMT14} & en $\to$ de & 24.53 & 24.50\phantom{*} & 25.04* & 24.74\phantom{*} & 22.73\phantom{*} & 25.00* & 24.57\phantom{*} \\
              & de $\to$ en & 28.01 & 28.65* & 28.83* & 29.39* & 28.24\phantom{*} & 29.13* & 29.97* \\
    & en $\to$ cs & 25.25 & 25.54\phantom{*} & 25.41\phantom{*} & 25.26\phantom{*} & 24.88\phantom{*} & 25.49\phantom{*} & 25.38\phantom{*} \\
    & cs $\to$ en & 28.78 & 28.84\phantom{*} & 29.64* & 29.41* & 25.77\phantom{*} & 29.23* & 29.15* \\
\hline
    \end{tabular}
  \end{center}
 \vspace*{-4mm}
   \caption{Main Results (BLEU(\%)) {\footnotesize($l$: sampling size in SR,
       $\alpha$: smoothing parameter).}
     \small{* indicates statistically significant difference ($p < 0.05$) from baselines
     with bootstrap resampling \cite{koehn2004statistical}. The same mark is used in Table 4 and 6.}}
   \label{result}
      \vspace*{-3mm}
\end{table*}

\subsection{Results with out-of-domain corpus}
To see the effect of subword regularization on a more open-domain
setting, we evaluate the systems with out-of-domain in-house data
consisting of multiple genres: Web, patents and query logs.  Note that we
did not conduct the comparison with KFTT and ASPEC corpora, as we found
that the domains of these corpora are too specific\footnote{KFTT focuses
on Wikipedia articles related to Kyoto, and ASPEC is a corpus of
scientific paper domain. Therefore, it is hard to translate
out-of-domain texts.}, and preliminary evaluations showed extremely poor
BLEU scores (less than 5) on out-of-domain corpora.

Table \ref{result2} shows the results. Compared to the gains obtained
with the standard in-domain evaluations in Table \ref{result}, subword
regularization achieves significantly larger improvements (+2 points)
in every domain of corpus. An interesting observation is that we have
the same level of improvements even on large training data sets (WMT14),
which showed marginal or small gains with the in-domain data. This
result strongly supports our claim that subword regularization is
more useful for open-domain settings.

  \begin{table}[ht]
      \renewcommand{\arraystretch}{0.9}
   \begin{center}
    \begin{tabular}[c]{l|l|l|c|c}
     \hline
     {\small\shortstack{Domain\\(size)}} & {\small Corpus} &
     {\small\shortstack{Language\\pair}} & {\small\shortstack{Baseline\\(BPE)}} &
     {\small\shortstack{Proposed\\(SR)}} \\
      \hline
     {\small Web} & {\small IWSLT15} & en $\to$ vi & 13.86 &  17.36* \\
     {\small (5k)} &      & vi $\to$ en & 7.83   & 11.69* \\
     &        & en $\to$ zh & 9.71 & 13.85* \\
     &       & zh $\to$ en & 5.93 & 8.13* \\
     &        {\small IWSLT17}  & en $\to$ fr & 16.09 &  20.04* \\
     &       & fr $\to$ en & 14.77 & 19.99* \\
      &       {\small WMT14} & en $\to$ de & 22.71 & 26.02* \\
      &       & de $\to$ en & 26.42 & 29.63* \\
      &       & en $\to$ cs & 19.53 & 21.41* \\
      &       & cs $\to$ en & 25.94 & 27.86* \\
      \hline
     {\small Patent} & {\small WMT14} & en $\to$ de & 15.63 & 25.76* \\
     {\small (2k)}   &                      & de $\to$ en & 22.74 & 32.66* \\
      &                           & en $\to$ cs & 16.70 & 19.38* \\
      &                           & cs $\to$ en & 23.20 & 25.30* \\
      \hline
     {\small Query} & {\small IWSLT15} & en $\to$ zh & 9.30 & 12.47* \\
     {\small (2k)} &                                 & zh $\to$ en & 14.94 & 19.99* \\
     &                {\small IWSLT17} & en $\to$ fr & 10.79 & 10.99\phantom{*} \\
     &                                 & fr $\to$ en & 19.01 & 23.96* \\
     & {\small WMT14} & en $\to$ de & 25.93 & 29.82* \\
                            &       & de $\to$ en & 26.24 & 30.90* \\
      \hline
    \end{tabular}
   \end{center}
   \vspace*{-4mm}
   \caption{Results with out-of-domain corpus \\
     \small{($l=\infty,\,\,\alpha=0.2$: IWSLT15/17,\,\,\,$l=64,\,\,\alpha=0.1$: others,\,\, one-best decding)}}
  \label{result2}
  \end{table}

\subsection{Comparison with other segmentation algorithms}
Table \ref{result3} shows the comparison on different segmentation
algorithms: word, character, mixed word/character \cite{wu2016google},
BPE \cite{sennrichneural} and our unigram model
with or without subword regularization.  The BLEU scores of word,
character and mixed word/character models are cited from
\cite{wu2016google}. As German is a morphologically rich language and
needs a huge vocabulary for word models, subword-based algorithms
perform a gain of more than 1 BLEU point than word model. Among
subword-based algorithms, the unigram language model with subword
regularization achieved the best BLEU score (25.04), which
demonstrates the effectiveness of multiple subword segmentations.

 \begin{table}[t]
     \renewcommand{\arraystretch}{0.9}
  \begin{center}
    \begin{tabular}[c]{l|l}
     \hline
     Model & BLEU \\
     \hline
     Word                     & 23.12 \\
     Character (512 nodes)    & 22.62 \\
     Mixed Word/Character     & 24.17 \\
     BPE                      & 24.53 \\
     \hline
     Unigram w/o SR ($l=1$)            & 24.50 \\
     Unigram w/ SR ($l=64,\,\,\alpha=0.1$) & 25.04 \\
      \hline
    \end{tabular}
  \end{center}
 \vspace*{-4mm}
 \caption{Comparison of different segmentation algorithms (WMT14 en$\rightarrow$de)}\label{result3}
 \vspace*{-3mm}
 \end{table}

\subsection{Impact of sampling hyperparameters}
\begin{figure}[t]
 \begin{center}
  \vspace*{-18mm}
 \includegraphics[width=110mm]{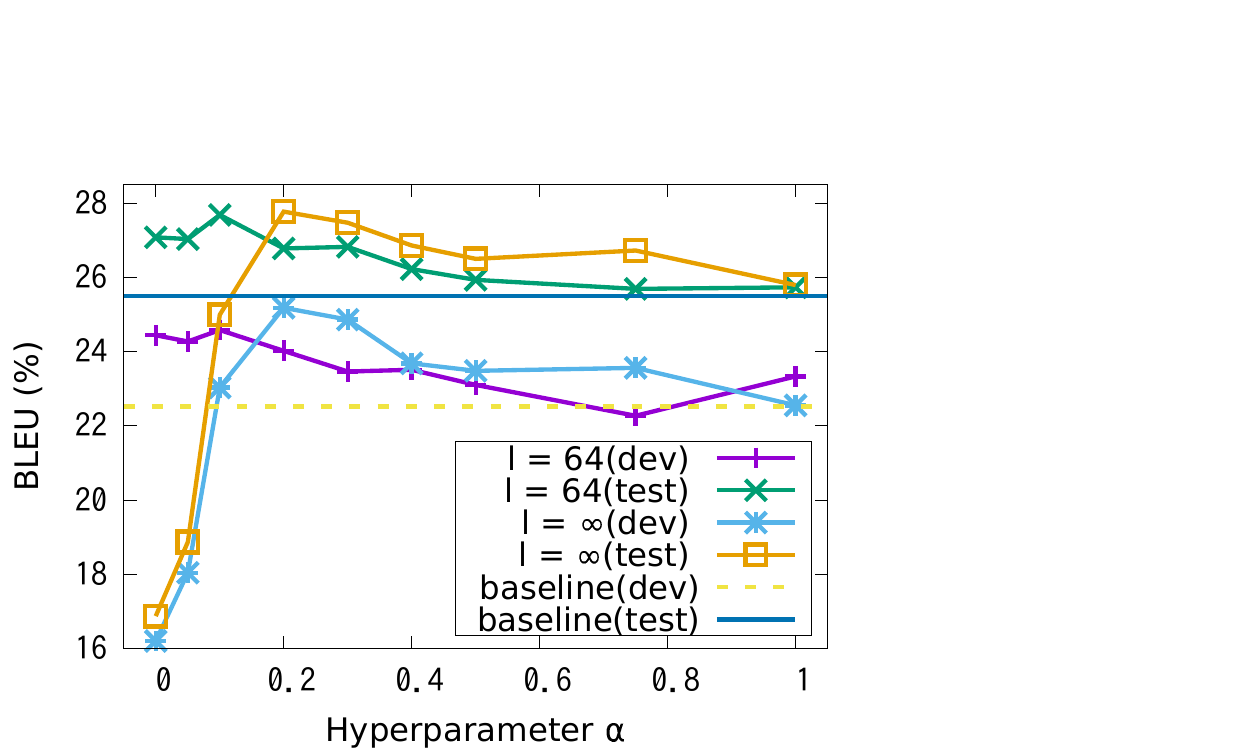}
  \vspace*{-10mm}
  \caption{̤Effect of sampling hyperparameters}\label{hyperparameter}
 \end{center}
\end{figure}

Subword regularization has two hyperparameters: $l$: size of sampling
candidates, $\alpha$: smoothing constant. Figure \ref{hyperparameter}
shows the BLEU scores of various hyperparameters on IWSLT15
(en $\rightarrow$ vi) dataset.

First, we can find that the peaks of BLEU scores against smoothing
parameter $\alpha$ are different depending on the sampling size $l$.
This is expected, because $l=\infty$  has larger search space than
$l=64$, and needs to set $\alpha$ larger to sample sequences close to
the Viterbi sequence $\mathbf{x}^{*}$.

Another interesting observation is that $\alpha=0.0$ leads to
performance drops especially on $l=\infty$. When $\alpha=0.0$, the
segmentation probability $P(\mathbf{x}|X)$ is virtually ignored and one
segmentation is uniformly sampled. This result suggests that biased
sampling with a language model is helpful to emulate the real noise in
the actual translation.

In general, larger $l$ allows a more aggressive regularization and is
more effective for low resource settings such as IWSLT. However, the
estimation of $\alpha$ is more sensitive and performance becomes even
worse than baseline when $\alpha$ is extremely small. To weaken the
effect of regularization and avoid selecting invalid parameters, it
might be more reasonable to use $l=64$ for high resource languages.

Although we can see in
general that the optimal hyperparameters are roughly predicted with
the held-out estimation, it is still an open question
how to choose the optimal size $l$ in subword sampling.

\subsection{Results with single side regularization}
Table \ref{result5} summarizes the BLEU scores with subword
regularization either on source or target sentence to figure out which
components (encoder or decoder) are more affected. As expected, we can
see that the BLEU scores with single side regularization are worse than
full regularization. However, it should be noted that single side
regularization still has positive effects. This result implies that
subword regularization is not only helpful for encoder-decoder
architectures, but applicable to other NLP tasks that only use an either
encoder or decoder, including text classification \cite{iyyer2015deep}
and image caption generation \cite{vinyals2015}.

 \begin{table}[t]
     \renewcommand{\arraystretch}{0.9}
  \begin{center}
    \begin{tabular}[c]{l|Q|Q|Q|Q}
     \hline
     {\small Regularization type} & en$\rightarrow$vi & vi$\rightarrow$en & en$\rightarrow$ar & ar$\rightarrow$en \\
     \hline
     {\small No reg. (baseline)} & 25.49\phantom{*} & 22.32\phantom{*} & 13.04\phantom{*} & 27.09\phantom{*} \\
     {\small Source only} & 26.00\phantom{*} & 23.09* & 13.46\phantom{*} & 28.16* \\
     {\small Target only} & 26.10\phantom{*} & 23.62* & 14.34* & 27.89*\\
     {\small Source and target} & 27.68* & 24.73* & 14.92* & 28.47* \\
      \hline
    \end{tabular}
  \end{center}
 \vspace*{-4mm}
 \caption{Comparison on different regularization strategies (IWSLT15/17,
     \,\,$l=64,\,\alpha=0.1$)}\label{result5}
 \vspace*{-4mm}
 \end{table}

\section{Conclusions}
In this paper, we presented a simple regularization method,
\textbf{subword regularization}\footnote{Implementation is available at
\url{https://github.com/google/sentencepiece}}, for NMT, with no change to the network
architecture.  The central idea is to virtually
augment training data with on-the-fly subword sampling, which helps to
improve the accuracy as well as robustness of NMT models.  In addition,
for better subword sampling, we propose a new subword segmentation
algorithm based on the unigram language model. Experiments on multiple
corpora with different sizes and languages show that subword
regularization leads to significant improvements especially on low
resource and open-domain settings.

Promising avenues for future work are to apply subword regularization to
other NLP tasks based on encoder-decoder architectures, e.g., dialog
generation \cite{vinyals20152} and automatic summarization
\cite{rush2015neural}. Compared to machine translation, these tasks do
not have enough training data, and thus there could be a large room for
improvement with subword regularization. Additionally, we would like to
explore the application of subword regularization for machine learning,
 including Denoising Auto Encoder
\cite{vincent2008extracting} and Adversarial Training
\cite{goodfellow2015}.

\bibliography{main} \bibliographystyle{acl_natbib}

\begin{thebibliography}{}
\expandafter\ifx\csname natexlab\endcsname\relax\def\natexlab#1{#1}\fi

\bibitem[{Artetxe et~al.(2017)Artetxe, Labaka, Agirre, and Cho}]{mikel2017}
Mikel Artetxe, Gorka Labaka, Eneko Agirre, and Kyunghyun Cho. 2017.
\newblock Unsupervised neural machine translation.
\newblock {\em arXive preprint arXiv:1710.11041\/} .

\bibitem[{Bahdanau et~al.(2014)Bahdanau, Cho, and Bengio}]{bahdanau2014neural}
Dzmitry Bahdanau, Kyunghyun Cho, and Yoshua Bengio. 2014.
\newblock Neural machine translation by jointly learning to align and
  translate.
\newblock {\em arXiv preprint arXiv:1409.0473\/} .

\bibitem[{Belinkov and Bisk(2017)}]{belinkov2017}
Yonatan Belinkov and Yonatan Bisk. 2017.
\newblock Synthetic and natural noise both break neural machine translation.
\newblock {\em arXive preprint arXiv:1711.02173\/} .

\bibitem[{Chan et~al.(2016)Chan, Zhang, Le, and Jaitly}]{chan2016latent}
William Chan, Yu~Zhang, Quoc Le, and Navdeep Jaitly. 2016.
\newblock Latent sequence decompositions.
\newblock {\em arXiv preprint arXiv:1610.03035\/} .

\bibitem[{Chitnis and DeNero(2015)}]{chitnis2015variable}
Rohan Chitnis and John DeNero. 2015.
\newblock Variable-length word encodings for neural translation models.
\newblock In {\em Proc. of EMNLP\/}. pages 2088--2093.

\bibitem[{Denkowski and Neubig(2017)}]{denkowski2017stronger}
Michael Denkowski and Graham Neubig. 2017.
\newblock Stronger baselines for trustable results in neural machine
  translation.
\newblock {\em Proc. of Workshop on Neural Machine Translation\/} .

\bibitem[{Gage(1994)}]{gage1994}
Philip Gage. 1994.
\newblock A new algorithm for data compression.
\newblock {\em C Users J.\/} 12(2):23--38.

\bibitem[{Gehring et~al.(2017)Gehring, Auli, Grangier, Yarats, and
  Dauphin}]{gehring2017convolutional}
Jonas Gehring, Michael Auli, David Grangier, Denis Yarats, and Yann~N Dauphin.
  2017.
\newblock Convolutional sequence to sequence learning.
\newblock {\em arXiv preprint arXiv:1705.03122\/} .

\bibitem[{Goodfellow et~al.(2015)Goodfellow, Shlens, and
  Szegedy}]{goodfellow2015}
Ian Goodfellow, Jonathon Shlens, and Christian Szegedy. 2015.
\newblock Explaining and harnessing adversarial examples.
\newblock In {\em Proc. of ICLR\/}.

\bibitem[{Iyyer et~al.(2015)Iyyer, Manjunatha, Boyd-Graber, and
  Daum{\'e}~III}]{iyyer2015deep}
Mohit Iyyer, Varun Manjunatha, Jordan Boyd-Graber, and Hal Daum{\'e}~III. 2015.
\newblock Deep unordered composition rivals syntactic methods for text
  classification.
\newblock In {\em Proc. of ACL\/}.

\bibitem[{Kingma and Adam(2014)}]{kingma6980method}
Diederik~P Kingma and Jimmy~Ba Adam. 2014.
\newblock A method for stochastic optimization.
\newblock {\em arXiv preprint arXiv:1412.6980\/} .

\bibitem[{Koehn(2004)}]{koehn2004statistical}
Philipp Koehn. 2004.
\newblock Statistical significance tests for machine translation evaluation.
\newblock In {\em Proc. of EMNLP\/}.

\bibitem[{Lample et~al.(2017)Lample, Denoyer, and Ranzato}]{guillaume2017}
Guillaume Lample, Ludovic Denoyer, and Marc'Aurelio Ranzato. 2017.
\newblock Unsupervised machine translation using monolingual corpora only.
\newblock {\em arXive preprint arXiv:1711.00043\/} .

\bibitem[{Luong et~al.(2015)Luong, Pham, and Manning}]{luong2015effective}
Minh-Thang Luong, Hieu Pham, and Christopher~D Manning. 2015.
\newblock Effective approaches to attention-based neural machine translation.
\newblock In {\em Proc of EMNLP\/}.

\bibitem[{Nagata(1994)}]{nagata1994stochastic}
Masaaki Nagata. 1994.
\newblock A stochastic japanese morphological analyzer using a forward-dp
  backward-a* n-best search algorithm.
\newblock In {\em Proc. of COLING\/}.

\bibitem[{Nakazawa et~al.(2017)Nakazawa, Higashiyama, Ding, Mino, Goto, Kazawa,
  Oda, Neubig, and Kurohashi}]{nakazawa2017overview}
Toshiaki Nakazawa, Shohei Higashiyama, Chenchen Ding, Hideya Mino, Isao Goto,
  Hideto Kazawa, Yusuke Oda, Graham Neubig, and Sadao Kurohashi. 2017.
\newblock Overview of the 4th workshop on asian translation.
\newblock In {\em Proceedings of the 4th Workshop on Asian Translation
  (WAT2017)\/}. pages 1--54.

\bibitem[{Nong et~al.(2009)Nong, Zhang, and Chan}]{nong2009linear}
Ge~Nong, Sen Zhang, and Wai~Hong Chan. 2009.
\newblock Linear suffix array construction by almost pure induced-sorting.
\newblock In {\em Proc. of DCC\/}.

\bibitem[{Papineni et~al.(2002)Papineni, Roukos, Ward, and
  Zhu}]{papineni2002bleu}
Kishore Papineni, Salim Roukos, Todd Ward, and Wei-Jing Zhu. 2002.
\newblock Bleu: a method for automatic evaluation of machine translation.
\newblock In {\em Proc. of ACL\/}.

\bibitem[{Rush et~al.(2015)Rush, Chopra, and Weston}]{rush2015neural}
Alexander~M Rush, Sumit Chopra, and Jason Weston. 2015.
\newblock A neural attention model for abstractive sentence summarization.
\newblock In {\em Proc. of EMNLP\/}.

\bibitem[{Schuster and Nakajima(2012)}]{schuster2012japanese}
Mike Schuster and Kaisuke Nakajima. 2012.
\newblock Japanese and korean voice search.
\newblock In {\em Proc. of ICASSP\/}.

\bibitem[{Scott(2002)}]{scott2002bayesian}
Steven~L Scott. 2002.
\newblock Bayesian methods for hidden markov models: Recursive computing in the
  21st century.
\newblock {\em Journal of the American Statistical Association\/} .

\bibitem[{Sennrich et~al.(2016)Sennrich, Haddow, and Birch}]{sennrichneural}
Rico Sennrich, Barry Haddow, and Alexandra Birch. 2016.
\newblock Neural machine translation of rare words with subword units.
\newblock In {\em Proc. of ACL\/}.

\bibitem[{Sperber et~al.(2017)Sperber, Neubig, Niehues, and
  Waibel}]{sperber2017neural}
Matthias Sperber, Graham Neubig, Jan Niehues, and Alex Waibel. 2017.
\newblock Neural lattice-to-sequence models for uncertain inputs.
\newblock In {\em Proc. of EMNLP\/}.

\bibitem[{Srivastava et~al.(2014)Srivastava, Hinton, Krizhevsky, Sutskever, and
  Salakhutdinov}]{srivastava2014dropout}
Nitish Srivastava, Geoffrey~E Hinton, Alex Krizhevsky, Ilya Sutskever, and
  Ruslan Salakhutdinov. 2014.
\newblock Dropout: a simple way to prevent neural networks from overfitting.
\newblock {\em JMLR\/} 15(1).

\bibitem[{Su et~al.(2017)Su, Tan, Xiong, Ji, Shi, and Liu}]{su2017lattice}
Jinsong Su, Zhixing Tan, De~yi Xiong, Rongrong Ji, Xiaodong Shi, and Yang Liu.
  2017.
\newblock Lattice-based recurrent neural network encoders for neural machine
  translation.
\newblock In {\em AAAI\/}. pages 3302--3308.

\bibitem[{Tai et~al.(2015)Tai, Socher, and Manning}]{tai2015improved}
Kai~Sheng Tai, Richard Socher, and Christopher~D Manning. 2015.
\newblock Improved semantic representations from tree-structured long
  short-term memory networks.
\newblock {\em Proc. of ACL\/} .

\bibitem[{Vaswani et~al.(2017)Vaswani, Shazeer, Parmar, Uszkoreit, Jones,
  Gomez, Kaiser, and Polosukhin}]{ashish2017google}
Ashish Vaswani, Noam Shazeer, Niki Parmar, Jakob Uszkoreit, Llion Jones,
  Aidan~N. Gomez, Lukasz Kaiser, and Illia Polosukhin. 2017.
\newblock Attention is all you need.
\newblock {\em arXive preprint arXiv:1706.03762\/} .

\bibitem[{Vincent et~al.(2008)Vincent, Larochelle, Bengio, and
  Manzagol}]{vincent2008extracting}
Pascal Vincent, Hugo Larochelle, Yoshua Bengio, and Pierre-Antoine Manzagol.
  2008.
\newblock Extracting and composing robust features with denoising autoencoders.
\newblock In {\em Proc. of ICML\/}.

\bibitem[{Vinyals and Le(2015)}]{vinyals20152}
Oriol Vinyals and Quoc~V. Le. 2015.
\newblock A neural conversational model.
\newblock In {\em ICML Deep Learning Workshop\/}.

\bibitem[{Vinyals et~al.(2015)Vinyals, Toshev, Bengio, and Erhan}]{vinyals2015}
Oriol Vinyals, Alexander Toshev, Samy Bengio, and Dumitru Erhan. 2015.
\newblock Show and tell: A neural image caption generator.
\newblock In {\em Computer Vision and Pattern Recognition\/}.

\bibitem[{Viterbi(1967)}]{viterbi67}
Andrew Viterbi. 1967.
\newblock Error bounds for convolutional codes and an asymptotically optimum
  decoding algorithm.
\newblock {\em IEEE transactions on Information Theory\/} 13(2):260--269.

\bibitem[{Wu et~al.(2016)Wu, Schuster et~al.}]{wu2016google}
Yonghui Wu, Mike Schuster, et~al. 2016.
\newblock Google's neural machine translation system: Bridging the gap between
  human and machine translation.
\newblock {\em arXiv preprint arXiv:1609.08144\/} .

\bibitem[{Xie et~al.(2017)Xie, Wang, Li, L{\'{e}}vy, Nie, Jurafsky, and
  Ng}]{xie2017}
Ziang Xie, Sida~I. Wang, Jiwei Li, Daniel L{\'{e}}vy, Aiming Nie, Dan Jurafsky,
  and Andrew~Y. Ng. 2017.
\newblock Data noising as smoothing in neural network language models.
\newblock In {\em Proc. of ICLR\/}.

\end{thebibliography}

\end{document}